\documentclass{article}

\usepackage[preprint]{neurips_2022}

\usepackage[utf8]{inputenc} % allow utf-8 input
\usepackage[T1]{fontenc}    % use 8-bit T1 fonts
\usepackage{hyperref}       % hyperlinks
\usepackage{url}            % simple URL typesetting
\usepackage{booktabs}       % professional-quality tables
\usepackage{amsfonts}       % blackboard math symbols
\usepackage{nicefrac}       % compact symbols for 1/2, etc.
\usepackage{microtype}      % microtypography
\usepackage{xcolor}         % colors
\usepackage{microtype}
\usepackage{wrapfig}
\usepackage{graphicx}
\usepackage{csquotes}
\usepackage{booktabs} % for professional tables
\usepackage{comment}
\usepackage{float}
\usepackage{makecell}
\usepackage{subcaption}
\usepackage{amsmath,amsfonts,bm}
\usepackage{booktabs}       % professional-quality tables

\newcommand{\privileged}{\mathcal{D}_{\mathrm{PI}}}

\newcommand{\searchspace}{\mathcal{E}}

\def\eqref#1{equation~\ref{#1}}
\def\1{\bm{1}}

\DeclareMathAlphabet{\mathsfit}{\encodingdefault}{\sfdefault}{m}{sl}
\SetMathAlphabet{\mathsfit}{bold}{\encodingdefault}{\sfdefault}{bx}{n}

\usepackage{caption} 
\usepackage{enumitem}% http://ctan.org/pkg/enumitem

\title{Controllable Neural Symbolic Regression}

\author{%
  Tommaso Bendinelli*\\
  CSEM SA, Alpnach\\
  \texttt{tommaso.bendinelli@csem.ch} \\
  \And
  Luca Biggio*\\
  ETH Z\"urich\\
  \texttt{luca.biggio@inf.ethz.ch } \\
  \And
  Pierre-Alexandre Kamienny\\
  Meta AI \& Sorbonne Université, CNRS, ISIR\\
  \texttt{pakamienny@meta.com}\\
}

\begin{document}

\maketitle

\begin{abstract}
In symbolic regression, the goal is to find an analytical expression that accurately fits experimental data with the minimal use of mathematical symbols such as operators, variables, and constants. However, the combinatorial space of possible expressions can make it challenging for traditional evolutionary algorithms to find the correct expression in a reasonable amount of time. To address this issue, Neural Symbolic Regression (NSR) algorithms have been developed that can quickly identify patterns in the data and generate analytical expressions. However, these methods, in their current form, lack the capability to incorporate user-defined prior knowledge, which is often required in natural sciences and engineering fields. To overcome this limitation, we propose a novel neural symbolic regression method, named Neural Symbolic Regression with Hypothesis (NSRwH) that enables the explicit incorporation of assumptions about the expected structure of the ground-truth expression into the prediction process. Our experiments demonstrate that the proposed conditioned deep learning model outperforms its unconditioned counterparts in terms of accuracy while also providing control over the predicted expression structure.
\end{abstract}

\section{Introduction}\label{sec:intro}
Symbolic Regression (SR) is a method that searches over the space of analytical expressions $\searchspace$ to find the best fit for experimental data by balancing the need to minimize expression complexity with maximizing accuracy. Unlike over-parametrized methods such as decision trees and neural networks, SR produces human-readable expressions that can provide valuable insights in fields such as material science \citep{wang2019symbolic,kabliman2021application,ma2022evolving} and fundamental physics \citep{schmidt2009distilling,vaddireddy2020feature,sun2022symbolic,cranmer2020discovering,hernandez2019fast,udrescu2020ai}. The goal of SR is to gain a deeper understanding of the underlying mechanisms of physical systems rather than trying to fit the data exactly, which can be affected by measurement errors. SR accomplishes this by converting input numerical data into compact and low-complexity representations in the form of symbolic mathematical expressions. \\ 
Researchers in natural sciences frequently rely on prior knowledge and analogies to comprehend novel systems and predict their behavior. When studying specific physical phenomena, scientists might anticipate particular constants or symmetries to appear in the mathematical laws describing the data. For instance, in astrophysics, the gravitational constant has a significant impact on determining the scale of interactions between celestial bodies, while in fluid dynamics, the Reynolds number denotes the relative significance of inertial and viscous forces. Thus, it is crucial to prioritize expressions that contain such constants while employing symbolic regression techniques, as they conform better to the physics laws governing the data.  
Access to a part of the underlying ground-truth system equation is also a common assumption made in the system identification literature where the physical laws are known up to a few parameters \citep{brunton2016discovering,kaheman2020sindy}. In our work, we will refer to the assumptions made by the SR practitioner about the underlying symbolic expression as \emph{hypotheses}. These hypotheses may be incomplete or partially incorrect and can be used in any form to restrict the search space. If a hypothesis is true, we will name it \textit{privileged information}.

\subsection*{Related work and background}
\paragraph{Genetic Programming.} Searching for a satisfactory analytical expression is a hard optimization problem, traditionally tackled using genetic programming (GP) algorithms. These methods work by i) defining a class of programs, represented in SR as tree structures where nodes are unary (e.g. $\texttt{cos},  \texttt{exp}$) or binary operators (e.g. $\texttt{add}, \texttt{mul}$) and leaves are variables and constants (e.g. $x_1$, 3.14) and ii) evolving a population of analytical expressions through selection, mutation, and crossovers. Being a greedy search approach, GP algorithms are prone to falling into local minima, and extensive exploration leads to relatively large run times. In practice with time constraints, such as the $24$ hours-limit in \cite{lacava}, the most accurate GP methods provide expressions with overly large complexity thus preventing the derivation of meaningful physical insights; on the Feynman datasets \citep{udrescu2020ai}, whose expressions have averaged complexity $20$ as defined in \cite{lacava}, the current  state-of-the-art \citep{10.1145/3377929.3398099} predicts expressions with averaged complexity $100$. Up to our knowledge, injection of prior information in GP methods can only be accomplished by filtering during selection, e.g. using properties like function positivity or convexity \citep{kronberger2022shape,haider2022comparing}. This strategy is inherently greedy and can result in the selection of suboptimal expressions due to early convergence to local minima. Other forms of high-level prior information available to the user, e.g. complexity of the expected expression, can hardly be incorporated into GP algorithms. Recently \citep{mundhenk2021symbolic}, a combination of neural networks and genetic programming (GP) has been proposed to improve the performance of symbolic regression. The neural network is used to generate the initial population for a GP algorithm, resulting in a hybrid approach that combines the strengths of both methods. This combination allows for the ability to learn patterns and explore a large solution space, resulting in remarkable performances. However, these systems are not easily controllable, meaning that it can be difficult for the user to constrain the predictions to conform to high-level properties that are known from prior knowledge of the problem. 
\vspace{-0.3cm}
\paragraph{AI-Feynman.} Recent studies \citep{udrescu2020ai,udrescu2} have investigated the idea of constraining the search to expressions that exhibit particular properties, such as compositionality, additivity, and generalized symmetry. By utilizing these properties, the task of SR becomes significantly less complex as it leverages the modular nature of the resulting expression trees. However, these approaches necessitate fitting a new neural network for every new input dataset and then examining the trained network to identify the desired properties, leading to an inevitably time-consuming process.
\vspace{-0.3cm}
\paragraph{Neural Symbolic Regression.} Inspired by recent advances in language models, a line of work named \textit{Neural Symbolic Regression} (NSR), tackles SR as a natural language processing task ~\citep{biggio2020seq2seq,biggio2021neural,valipour2021symbolicgpt,d2022deep,kamienny2022end, Vastl2022SymFormerES, Li2022SymbolicET, becker2022discovering}. 
NSR consists of two primary steps: firstly, large synthetic datasets are generated by i) sampling expressions from a prior distribution $p_\theta(\searchspace)$ where $\theta$ is a parametrization induced by an off-the-shelf expression generator \citep{lample2019deep}, ii) evaluating these expressions on a set of points $\mathbf{x}\in \mathbb{R}^d$ where $d$ is the feature dimension, e.g. sampled from a uniform distribution. Secondly, a generative model $g_\phi(\searchspace|\mathcal{D})$, practically a Transformer \citep{vaswani2017attention} parametrized by weights $\phi$, that is conditioned on input points $\mathcal{D}=(\mathbf{x}, \mathbf{y})$, is trained on the task of next-token prediction with target the Polish notation of the expression. NSR predicts expressions that share properties of their implicitly biased synthetic generator $p_\theta(\searchspace)$. Control over the shape of the predicted expressions, e.g complexity or sub-expression terms, boils down to a sound design of the generator and the pipeline introduced in \cite{lample2019deep} allows only limited degrees of freedom such as operators, variables, constants probability, and tree depth.\\
Similarly to querying a text-to-image generative model \citep{ramesh2022hierarchical,saharia2022photorealistic} with a prompt, the SR practitioner might want to restrict the class of predicted expressions to be in a subclass $h(\searchspace) \subset \searchspace$ by using privileged information.  Examples of $h(\searchspace)$ can be the class of expressions with low complexity, or that include a specific sub-expression like $e^{-\sqrt{x_1^2+x_2^2}}$. However, a trained NSR model $g_\phi(\searchspace|\mathcal{D})$ can only be adapted to $h(\searchspace)$ in one of two ways: i) by using rejection sampling, which is time-inefficient and does not guarantee to find candidate expressions with the expected inductive biases, or ii) by designing a new generator with the desired properties and fine-tuning the model on the new dataset, which is a tedious and time-consuming task.

\subsection*{Contributions}
In this work, we propose a new method called \emph{Neural Symbolic Regression with Hypotheses} (NSRwH) to address the aforementioned limitations of NSR algorithms. NSRwH efficiently restricts the class of predicted expressions of NSR models \emph{during inference}, if provided privileged information $\privileged$, with a simple modification to both the model architecture and the training data generation: with the training set of expressions from $p_\theta(\searchspace)$, we produce descriptions $\privileged$, e.g. appearing operators or complexity, and feed this meta-data into the Transformer model as an extra input, i.e. $g_\phi(S|\mathcal{D}, \privileged)$. During training, we use a masking strategy to avoid our model considering sub-classes of expressions when no privileged information is provided. We show that our model exhibits the following desirable characteristics:

\begin{enumerate}[leftmargin=*]
    \item In a similar vein to the recent literature on expression derivation and integration \citep{lample2019deep} and mathematical understanding capabilities of Transformers \citep{charton2022my}, our results demonstrate that Transformer models can succeed in capturing complex, high-level symbolic expression properties, such as complexity and symmetry. 
    \item The proposed model is able to output expressions that closely align with user-determined privileged information and/or hypotheses on the sought-for expression when it is conditioned on such information. This makes the model effectively \emph{controllable} as its output reflects the user's expectations of specific high-level properties. This stands in contrast to previous work in the NSR and GP literature, where steering symbolic regressors toward specific properties required either retraining from scratch or using inefficient post hoc greedy search routines.  
    \item The injection of privileged information provides significant improvements in terms of recovery rate. Such an improvement is, as expected, proportional to the amount of conditioning signal provided to the model. This effect is even more apparent in the case where numerical data are corrupted by noise and in the small data regime, where standard NSR approaches witness a more marked performance deterioration. 
    \item  We empirically demonstrate that incorporating conditioning hypotheses not only enhances the controllability of NSRwH but also improves its exploration capabilities, in contrast to standard NSR approaches that rely solely on increasing the beam size. In particular, we show that injecting a large number of hypotheses randomly chosen from a large pool of candidates results in better exploration performance compared to a standard NSR approach operating with a large beam size.
\end{enumerate}
 
In essence, our approach provides an additional degree of freedom to standard NSR algorithms by allowing the user to quickly steer the prediction of the model in the direction of their prior knowledge at inference time. This is accomplished by leveraging established techniques from language modeling and prompt engineering. The paper is structured as follows: in Section \ref{sec:method}, we describe our data generation pipeline and the model architecture; in Section \ref{sec:exps} we detail our experimental setup and report our empirical results and in Section \ref{sec:conc} we discuss promising future directions and the current limitations of our approach. 

\section{Method}
\label{sec:method}

\begin{figure}[ht!]
\centering
 \includegraphics[width=0.7\textwidth,height=5cm]{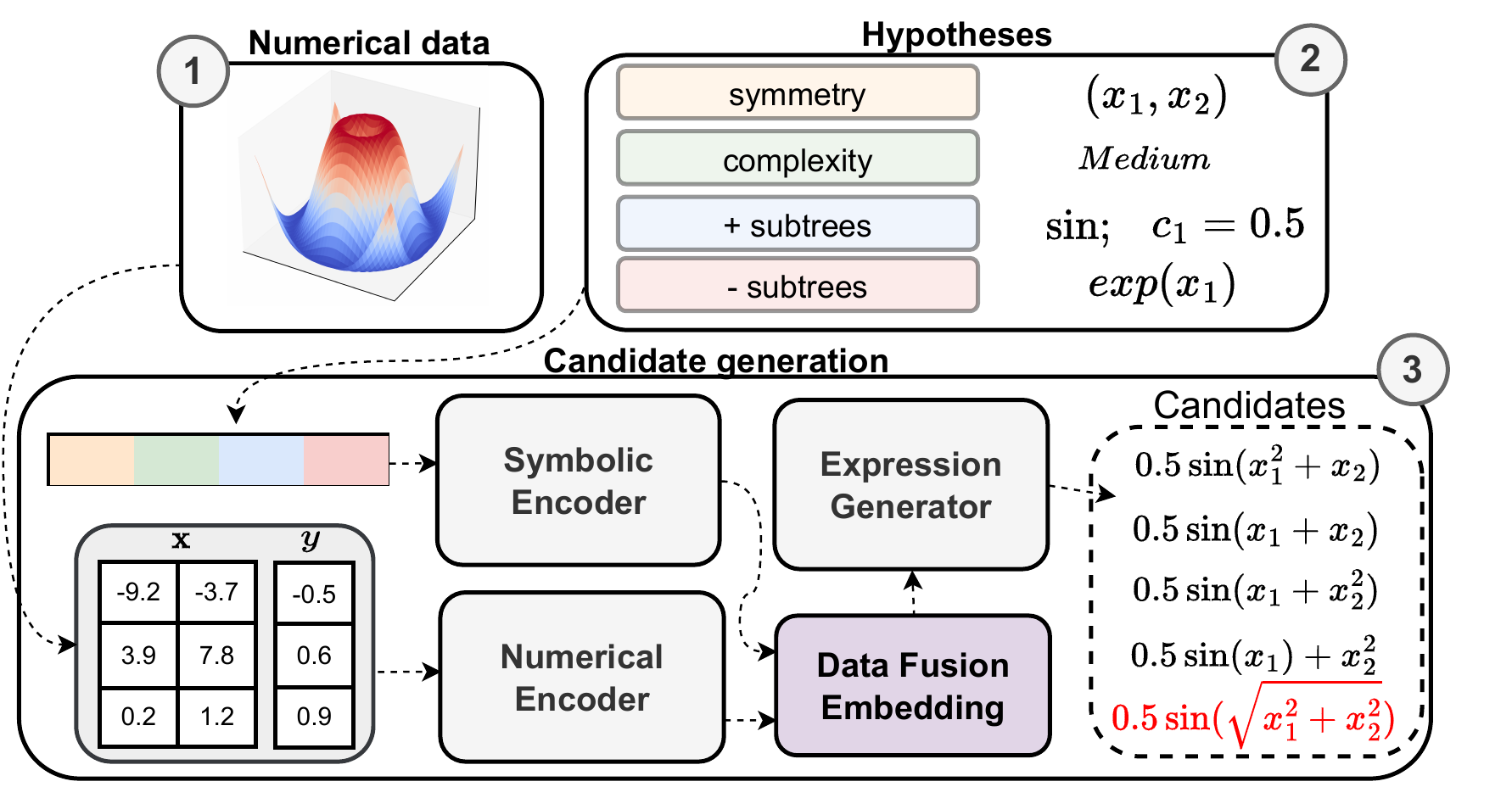}
\caption{\textbf{Neural Symbolic Regression with Hypotheses.} 1) A dataset of numerical observations is obtained; 2) the user formulates a set of hypotheses based on some properties they believe the final expression should possess. After being tokenized independently, the properties tensors are concatenated to form a unique conditioning tensor; 3) numerical data as well as the formulated hypotheses are given as input to two different encoders. Their outputs are then summed and the resulting tensor is processed by a decoder which outputs a set of candidate equations. For NSRwH to be effective and controllable, the candidate expressions should respect the input hypotheses.} 
\label{fig:model}
\end{figure}

\subsection{Notation and framework}
\label{subsec:notation}
A symbolic regressor is an algorithm that takes as input a dataset $\mathcal{D}$ of $n$ features-value pairs $(\mathbf{x_i}, y_i)\sim \mathbb{R}^d \times \mathbb{R}$, where $d$ is the feature dimension, and returns a symbolic expression $e\sim\mathcal{E}$ such that $\forall(\mathbf{x}_i, y_i) \in \mathcal{D}, \ e(\mathbf{x}_i) = \tilde{y_i} \approx y_i$. NSR is a class of SR algorithms that learns a distribution model $g_\phi(\mathcal{E} \mid \mathcal{D})$, parametrized by a neural network with weights $\phi$, over symbolic expressions conditioned on an input dataset $\mathcal{D}$.
In this work, we introduce NSRwH, a new subclass of neural symbolic regressors, that allows for conditioning their predictions with user-specified prior knowledge about the output expression. More concretely, given a set of privileged information $\privileged$, NSRwH approaches are trained to model the conditional distribution $g_\phi(\mathcal{E} \mid \mathcal{D},\privileged)$. An illustration of the proposed approach is shown in Fig. \ref{fig:model}.

\subsection{Dataset generation}
\label{subsec:data_generation}

In our framework, a synthetic training sample is defined as a tuple $(e, \mathcal{D},\privileged)$ where each element is produced as explained below.

\paragraph{Generating $e$ and $\mathcal{D}$.} As in other NSR works \citep{biggio2021neural, kamienny2022end, valipour2021symbolicgpt}, we sample analytical expressions $e$ from $\searchspace$ using the strategy introduced by ~\citet{lample2019deep}:  random unary-binary trees with depth between $1$ and $6$ are generated, then internal nodes are assigned either unary or binary operators as described in Table \ref{tab:operators} in Appendix \ref{app1_1} according to their arity, and leaves are assigned variables $\{x_d\}_{d \leq 5}$ and constants. 
In order to generate $\mathcal{D}$, for each expression $e$, we sample a support of $n$ points $\mathbf{x}_i \in \mathbb{R}^d$.
The values for each coordinate are drawn independently from one another using a uniform distribution $\mathcal{U}$, with the bounds randomly selected from the interval $[-10,10]$.  
Next, the expression value $y_i$ is obtained via the evaluation of the expression $e$ on the previously sampled support. More details on the generation of numerical data can be found in Appendix \ref{app1_1}

\paragraph{Generating $\privileged$.}  Privileged information $\privileged$ is composed of \textit{hypotheses}. From an expression $e$, we extract the following properties:

\begin{itemize}[leftmargin=*]
\item \textbf{Complexity.} We use the definition of complexity provided by \cite{lacava}, i.e. the number of mathematical operators, features, and constants in the output prediction. 
\item \textbf{Symmetry.} We use the definition of generalized symmetry proposed in \citep{udrescu2}: $f$ has generalized symmetry if the $d$ components of the vector $\mathbf{x} \in \mathbb{R}^d$ can be split into groups of $k$ and $d-k$ components (which we denote by the vectors $\mathbf{x}^{\prime} \in \mathbb{R}^k$ and $\mathbf{x}^{\prime \prime} \in \mathbb{R}^{d-k}$ ) such that $f(\mathbf{x})=f\left(\mathbf{x}^{\prime}, \mathbf{x}^{\prime \prime}\right)=g\left[h\left(\mathbf{x}^{\prime}\right), \mathbf{x}^{\prime \prime}\right]$ for some unknown function $g$.
 \item \textbf{Appearing branches.} 
We consider the set of all the branches that appear in the prefix tree of the generating expression.
For instance, for $x_1 + \sin x_2$ this set would be $[+, x_1, +x_1, \sin, \sin(x_2), x_2, + \sin, + \sin(x_2)]$. For each expression, in the training set, we sample a subset of this list, ensuring that each element of the subset is sampled with a probability inversely proportional to its length squared and that the full expression tree is never given to the model.
\item \textbf{Appearing constants.}
We also enable the inclusion of a-priori-known constants at test time. We implement this conditioning by drawing inspiration from the concept of pointers in computer programming: we give as input to the model the numerical constant and a pointer, and the model has to place the input pointer in the correct location in the output prediction. This approach does not require representing each constant with a different token, hence preventing the explosion of the output vocabulary size.
\item \textbf{Absent branches.}
We condition our model with the information about subtrees not appearing in true expression. The procedure for extracting this property follows the same logic as the extraction of appearing subtrees.
\end{itemize}

In the rest of the paper, we refer to these properties as \texttt{Complexity}, \texttt{Symmetry}, \texttt{Positive}, \texttt{Constants}, and \texttt{Negative}. It is important to note that the set of properties used in this work is not exhaustive and can easily be expanded based on the user's prior knowledge. We provide more details on their exact computation along with a practical example of their extraction in Appendix \ref{app1_2}.

\subsection{Model}
\label{subsec:model}

\paragraph{Architecture.}
We use NeSymReS \citep{biggio2021neural} as our base neural symbolic regressor for its simplicity and in the following, we explain how to incorporate the description $\privileged$ as an input to the model $g_\phi(e|\mathcal{D},\privileged)$. Note that the very same conditioning strategy can easily be applied to alternative more advanced NSR architectures, such as those introduced in \citep{valipour2021symbolicgpt,kamienny2022end}.
NSRwH consists of three architectural components: a numerical encoder $enc_{num}$, a symbolic encoder $enc_{sym}$, and a decoder $dec$ (see Fig. \ref{fig:model}). Numerical data $\mathcal{D}$, represented by a tensor of size $(B, n, D)$, where $B$ is the batch size, $n$ is the number of points and $D$ is the sum of dependent/independent variables ($D=5+1$), is converted into a higher dimensional tensor $\mathcal{D}'$
 of size $(B, n, H)$ using a multi-hot bit representation according to the half-precision IEEE-754
standard and an embedding layer, where H is the hidden dimension (512 for our experiments). $\mathcal{D}'$ is then processed by a set-transformer encoder \citep{lee2019set}, a variation of \citep{vaswani2017attention} with better inference time and less memory requirement, to produce a new tensor $\mathbf{z}_{num} = enc_{num}(\mathcal{D}')$ of size $(B, S, H)$, where $S$ (50 for our experiments) is the sequence length after the encoder processing. $\privileged$, represented by a tensor of size $(B, M)$ where $M$ is the number of tokens composing the conditioning hypotheses string, is converted into a higher dimensional tensor $\mathcal{D}_{PI}'$ of size $(B, M, H)$ via an embedding layer. This new tensor is then input into an additional set-transformer to produce a tensor $\mathbf{z}_{sym} = enc_{sym}(\mathcal{D}_{PI}')$ of size $(B, S, H)$. $\mathbf{z}_{num}$ and $\mathbf{z}_{sym}$ are summed together to produce a new tensor $\mathbf{z}_{fused} = \mathbf{z}_{num} + \mathbf{z}_{sym}$ of size $(B, S, H)$. Finally, $\mathbf{z}_{fused}$ is fed into a standard transformer decoder network, $dec$, that autoregressively predicts token by token the corresponding expressions using beam search for the best candidates. We resorted to the element-wise summation of $\mathbf{z}_{num}$ and $\mathbf{z}_{sym}$ instead of concatenation in order to reduce memory usage in the decoder, which increases quadratically with the sequence length due to cross attention.

\vspace{-0.3cm}

\paragraph{Training and testing.}
As done in all NSR approaches, we use the cross-entropy loss on next-token prediction using teacher-forcing \citep{sutskever2014sequence}, i.e. conditioning $g_\phi(\tilde{e}_{t+1} | e_{1:t}, \mathcal{D}, \privileged)$ on the first $t$ tokens of the ground-truth $e$. As for NeSymReS, we \enquote{skeletonize} target expressions by replacing constants by a constant token $\diamond$ or, in the case the position of the constant is known a priori, a pointer symbol is used. To prevent our model from being dependent on privileged information at test time, we include training examples with partial privileged information. This means we only provide the model with a subset of all the possible conditionings. For example, only \texttt{Positive} and \texttt{Symmetry} are given, while \texttt{Negative}, \texttt{Complexity} and \texttt{Constants} are masked out. This is a useful feature of our model as, depending on the use case, some information might not be available and we want the model to still be usable in those cases. At test time, as for NeSymReS, we use beam search to produce a set of predicted expressions, then we apply Broyden–Fletcher–Goldfarb–Shanno algorithm (BFGS) \cite{Flet87} to recover the values of the constants by minimizing the squared loss between the original outputs and the output from the predicted expressions. More details on the model and training hyperparameters can be found in Appendix \ref{app3}.
\vspace{-0.2cm}
\section{Experiments}\label{sec:exps} 
\vspace{-0.3cm}
In this section, we first introduce the datasets and metrics used to evaluate the model and then we present our experiments aimed to assess different properties of NSRwH, including its controllability, and its performance when $\privileged$ is available, and when it is not. Over the experimental section, we use the standard NeSymReS as a reference baseline, which is referred to as \texttt{standard\_nesy} in the plots. While our approach could be used with other NSR methods, we have chosen to solely focus on NeSymReS as a baseline model. This allows us to better comprehend the advantages that come from conditioning, instead of assessing various NSR models with distinct numerical input architectures and expression generators. As mentioned in Section \ref{sec:intro}, GP methods can be hardly conditioned on our set of properties, and as such a comparison with them would be unfair.

\subsection{Experimental setup}
To generate training data, we follow the pipeline introduced in Section \ref{subsec:data_generation} resulting in a training set comprising 200 million symbolic expressions with up to 5 variables. The datasets and metrics used to test NSRwH are described below.
\paragraph{Datasets.}
We use five different databases in our experiments, each characterized by different degrees of complexity: 1) \texttt{train$\_$nc}: this dataset comprises 300 symbolic expressions, not including numerical constants. The number of independent variables varies from 1 to 5. The equations are sampled from the same distribution of the training set; 2) \texttt{train$\_$wc}: it comprises the same equations of \texttt{train$\_$nc} but with numerical constants randomly included in each expression. As such, it represents a more challenging framework than the previous one as the model has the output constant placeholders in the correct positions and BFGS has to find their numerical value; 3) \texttt{only$\_$five$\_$variables$\_$nc}: it consists of 300 expressions without constants, strictly selected to have 5 independent variables each. The dataset has been chosen to assess the performance of our algorithm in a higher-dimensional scenario; 4) \texttt{AIF}: it comprises all the equations with up to 5 independent variables extracted from the publicly available AIFeynman database \citep{udrescu2020ai}. It includes equations from the \textit{Feynman Lectures on Physics series} and serves to test the performance of NSRwH on mathematical expressions stemming from several physics domains; 5) \texttt{black$\_$box}: it is extracted from the ODE-Strogatz \citep{strogatz2018nonlinear} databases and serves to evaluate NSRwH in the case where no prior information is available. As also noted by \citet{kamienny2022end}, these datasets are particularly challenging as they include non-uniformly distributed points and have different input support bounds than those used by our dataset generation pipeline.

\paragraph{Metrics.}
We use three different metrics to evaluate our models: 1) \texttt{is$\_$satisfied}: this metric measures the percentage of output predictions that agree with a certain property. For all the properties this metric is calculated as follows: given a known equation, we calculate the mean over the total number of times the predictions of the model across the beam size matches the property under consideration. The final metric value is given by the average of the above quantity across all the equations in the test set; 2) \texttt{is\_correct}: given a test equation, for each point $(x,y)$ and prediction $\hat{y}$, we calculate  \texttt{numpy.is\_close($y$,$\hat{y}$)}. Then, we take the mean over all the support points and obtain a real number. If this number is larger than 0.99, we deem our prediction to match the true one and we assign a score of 1, otherwise 0. The final metric value is obtained by calculating the percentage of correctly predicted equations over the entire test set. Importantly, the support points are chosen to be different from those fed into the model at test time; 3) \texttt{$R^2_{\text{mean}}$}: given a test equation, and $n$ points $\{x_i,y_i\}_{i=1}^n$, and the corresponding predictions $\{\hat{y}_i\}_{i=1}^n$, we calculate the coefficient of determination, also known as $R^2$ score, as defined below:
    \vspace{-0.2cm}
    \begin{equation*}
    R^2=1-\frac{\sum_{i=1}^n\left(y_i-\hat{y}_i\right)^2}{\sum_{i=1}^n\left(y_i-\bar{y}_i\right)^2} \hspace{0.5cm}  \text{where} \hspace{0.5cm} \bar{y} = \frac{1}{n}\sum_{i=1}^ny_i      
    \end{equation*}
The final metric is calculated by taking the mean of the $R^2$ scores obtained for each equation in the test set. More details on the test datasets and metrics can be found in Appendix \ref{app2}

\subsection{Can transformers efficiently restrict the inference
space using descriptions?} 
\begin{figure}[H]
\centering
\includegraphics[scale = 0.24]{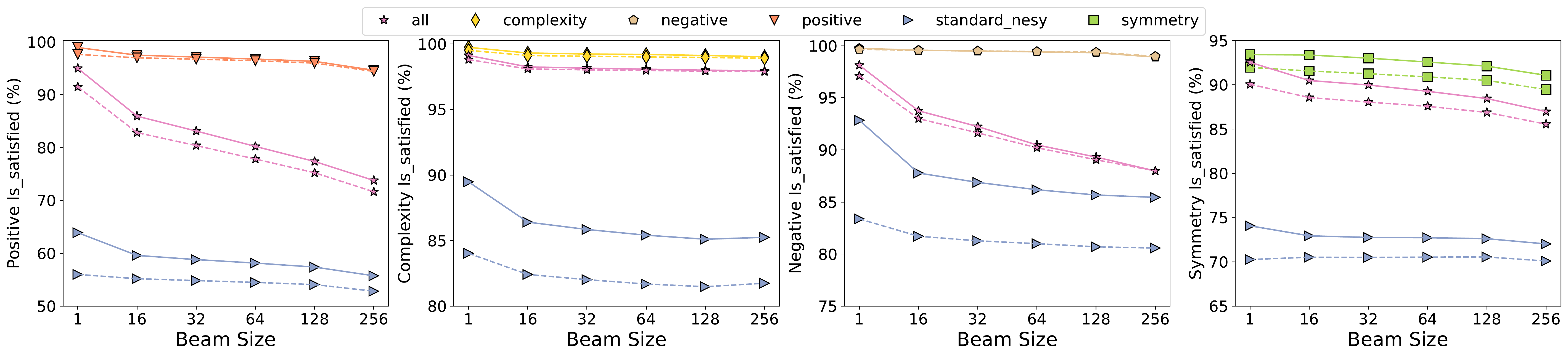} \\
\caption{\textbf{Controllability and property matching}: The panels show the level of agreement with various types of input conditioning signals -- in terms of the \texttt{is$\_$satisfied} metric -- of our model and the unconditioned baseline (\texttt{standard$\_$nesy}), both in the noiseless case (full line) and when noise is injected in the input data (dashed line), as a function of the beam size. The reported results are averaged across all datasets apart from \texttt{black$\_$box}.}
\label{property_matching}
\end{figure}

Arguably, the main challenge in symbolic regression is represented by the extremely large search space over mathematical expressions. Methods based on brute force search techniques are doomed to fail or to fall into spurious local minima. The goal of this section is to show that neural symbolic regression algorithms can be \emph{controlled} in such a way that their output adheres with a set of pre-specified inductive biases -- meant to narrow the search space -- on the nature of the sought expression. Each panel in Fig. \ref{property_matching} shows the evolution of the \texttt{is$\_$satisfied} metric for various types of conditioning properties as the beam size increases, with and without noise injected in the input data. Noise perturbations are injected in the output of the input data, $y$, according to the following formula:
\begin{equation}\label{noise injection}
\tilde{y} = y + \rho\epsilon \hspace{0.5cm}    \text{where} \hspace{0.5cm} \epsilon\sim\mathcal{N}(0,\vert{y}\vert) \hspace{0.2cm} and \hspace{0.2cm} \rho=0.01.
\end{equation}
The goal of the experiment is twofold: first, we want to assess whether NSRwH is able to capture the meaning of the input conditing, and second, we want to verify how consistent such an agreement is as we increase the beam size and inject noise. From the results in Fig. \ref{property_matching}, we can observe that the predictions of NSRwH attain a very high \texttt{is$\_$satisfied} score for all the evaluated properties. This is in contrast with the unconditioned model which does not consistently capture the underlying properties. This is particularly evident when noise is added to the data, as our model shows robustness to such perturbations, while the standard NSR method experiences greater variations. This is explained by the fact that the standard method grounds its predictions solely on numerical data. As such, when these are severely corrupted, results deteriorate accordingly.
We also note that when all possible conditioning properties are given to the model (see \texttt{all}), NSRwH tends to underperform with respect to the case when a single property is provided, in particular as the beam size increases. This is likely due to interference effects between different hypotheses, which causes the model, at large beam sizes, to select the subset of them that is more consistent with the numerical data.
\vspace{-0.3cm}

\vspace{-.7cm}
\subsection{Can NSRwH leverage privileged information?}\label{sec:privabailable}
\begin{wrapfigure}{r}{0.5\textwidth}
\vspace{-0.5cm}
  \begin{center}
    \includegraphics[width=0.45\textwidth]{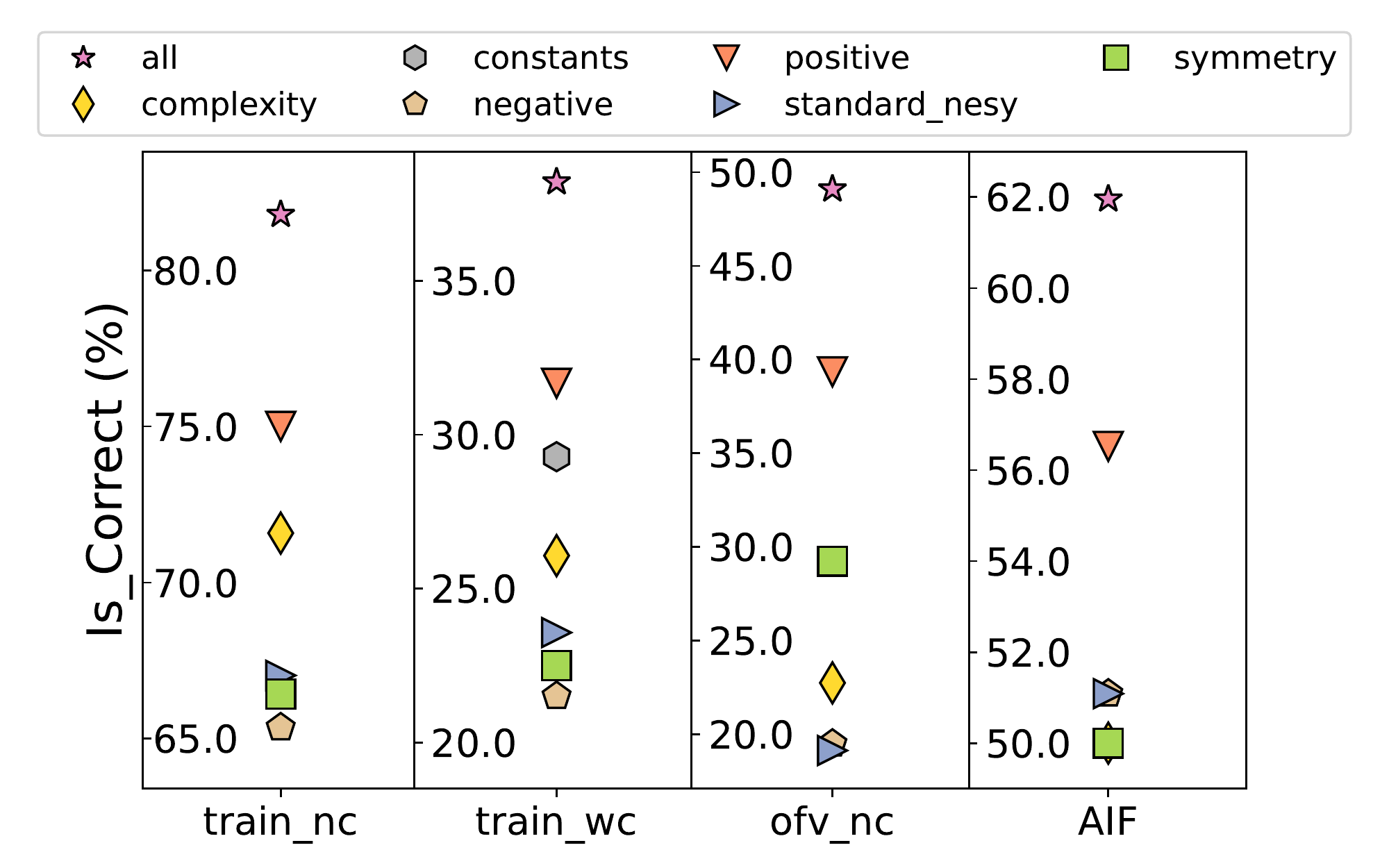}
  \end{center}
\caption{\textbf{Conditioning improves performance.} Comparison between NSRwH conditioned with different types of hypotheses and the unconditioned baseline (\texttt{standard$\_$nesy}) in terms of the \texttt{is$\_$correct} metric. Each column corresponds to a different test dataset.} \label{noiseless}
\end{wrapfigure}
In this section, we investigate whether the ability of NSRwH to capture the meaning of the input properties can be leveraged to improve performance. To perform these experiments we make use of \texttt{is$\_$correct} metric introduced above and we study how performance changes under the effect of noise, number of input data, and amount of conditioning. The beam size for both NSRwH and NeSymReS is set to 5. We start our evaluation from the noiseless case, i.e. no noise is injected in the expressions' evaluation at test time. As such, the mapping between input covariates and the output value is exactly represented by the ground truth symbolic expression. Fig. \ref{noiseless}, shows the performance of NSRwH and the unconditioned model in terms of the  \texttt{is$\_$correct} metric described above and the different properties provided at test time. Generally, NSRwH efficiently leverages the prompted information to improve its performance. Among the considered individual properties, \texttt{Positive} is  the most effective one. However, it is interesting to note that \texttt{Symmetry} is particularly effective on the \texttt{only\_five\_variables\_nc} (ofv$\_$nc) dataset. This is due to the high-dimensional nature of the dataset and the fact the symmetry information is more useful in such cases. Providing information about the ground-truth constants leads to significant performance improvements on the \texttt{train$\_$wc} dataset, showcasing the effectiveness of our strategy of providing numerical constants to the model. Finally, \texttt{all}, the combination of all the considered properties, is by far the most impacting conditioning. It is noteworthy that, while in some cases the performance of individual properties may not be significantly better than the baseline, their combination (\texttt{all}) proves to be highly successful, indicating that the model is able to combine them together effectively.

\subsubsection{Case with noise}

\begin{figure}[H]
\centering
\includegraphics[scale = 0.24]{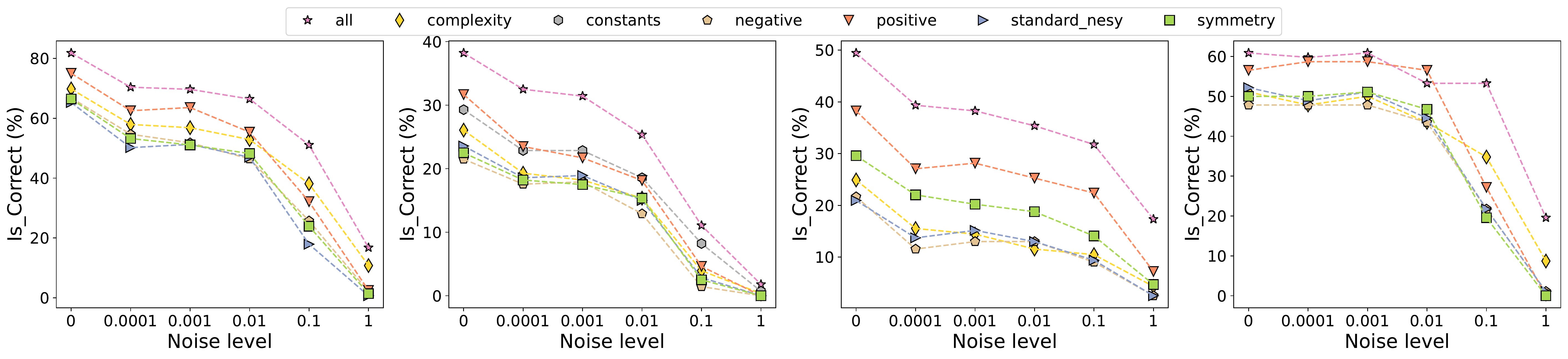}\\
\caption{\textbf{Dependence on the input noise.} Comparison between NSRwH conditioned with different types of hypotheses and the unconditioned baseline (\texttt{standard$\_$nesy}) in terms of the \texttt{is$\_$correct} metric, as a function of the noise level in the input data, for the 
\texttt{train$\_$nc},
\texttt{train$\_$wc}, \texttt{only$\_$five$\_$variables$\_$nc} and \texttt{AIF} datasets from left to right.}\label{noise}
\end{figure}

In this paragraph, we explore the more challenging scenario where noise is injected into the output value $y$ at test time. In particular, we use Eq. \ref{noise injection} with six different noise levels $\rho\in\{0,0.0001,0.001,0.01,0.1,1\}$. The beam size for both NSRwH and NeSymReS is set to 5 in this experiment. As shown in Fig. \ref{noise}, the performance improvements are even more pronounced than in the noiseless case shown in Fig. \ref{noiseless}. This illustrates that the incorporation of meaningful inductive biases in our model enables it to effectively manage the impact of noise and, as a result, improves generalization. 
\vspace{-0.4cm}

\vspace{-0.6cm}
\subsubsection{Dependence on the number of input points.}
\begin{figure}[H]
\centering
\includegraphics[scale = 0.24]{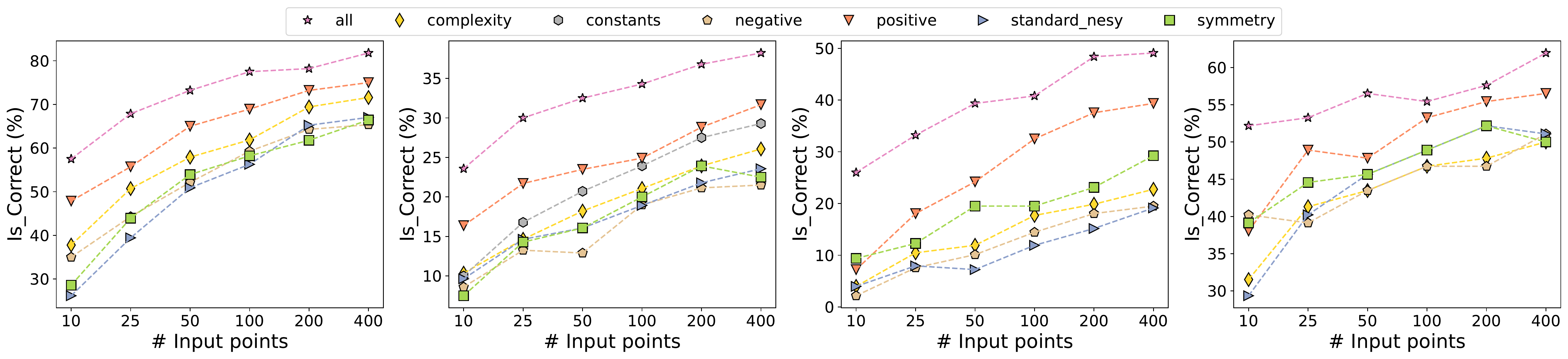}\\
\caption{\textbf{Dependence on the number of input points.} Comparison between NSRwH conditioned with different types of hypotheses and the unconditioned baseline (\texttt{standard$\_$nesy}) in terms of the \texttt{is$\_$correct} metric, as a function of the number of input points, for the \texttt{train$\_$nc},
\texttt{train$\_$wc}, \texttt{only$\_$five$\_$variables$\_$nc} and \texttt{AIF} datasets from left to right.}\label{input points}
\end{figure}

\vspace{-0.2cm}
In a similar manner as the previous paragraph, this investigation examines whether NSRwH can utilize input conditioning to enhance its performance in the challenging, yet common scenario where small datasets are used as input. As before, the beam size for both NSRwH and NeSymReS is set to 5. As illustrated in Fig. \ref{input points}, as the number of input points decreases, the performance of both the conditioned and unconditioned models also declines. However, in NSRwH this effect is significantly reduced, keeping relatively high levels of accuracy even when working in the small data regime.
\vspace{-0.2cm}

\vspace{-0.5cm}
\subsubsection{Dependence on the amount of conditioning}
\begin{wrapfigure}{r}{0.56\textwidth}
\vspace{-0.6cm}
  \begin{center}
    \includegraphics[width=0.33\textwidth]{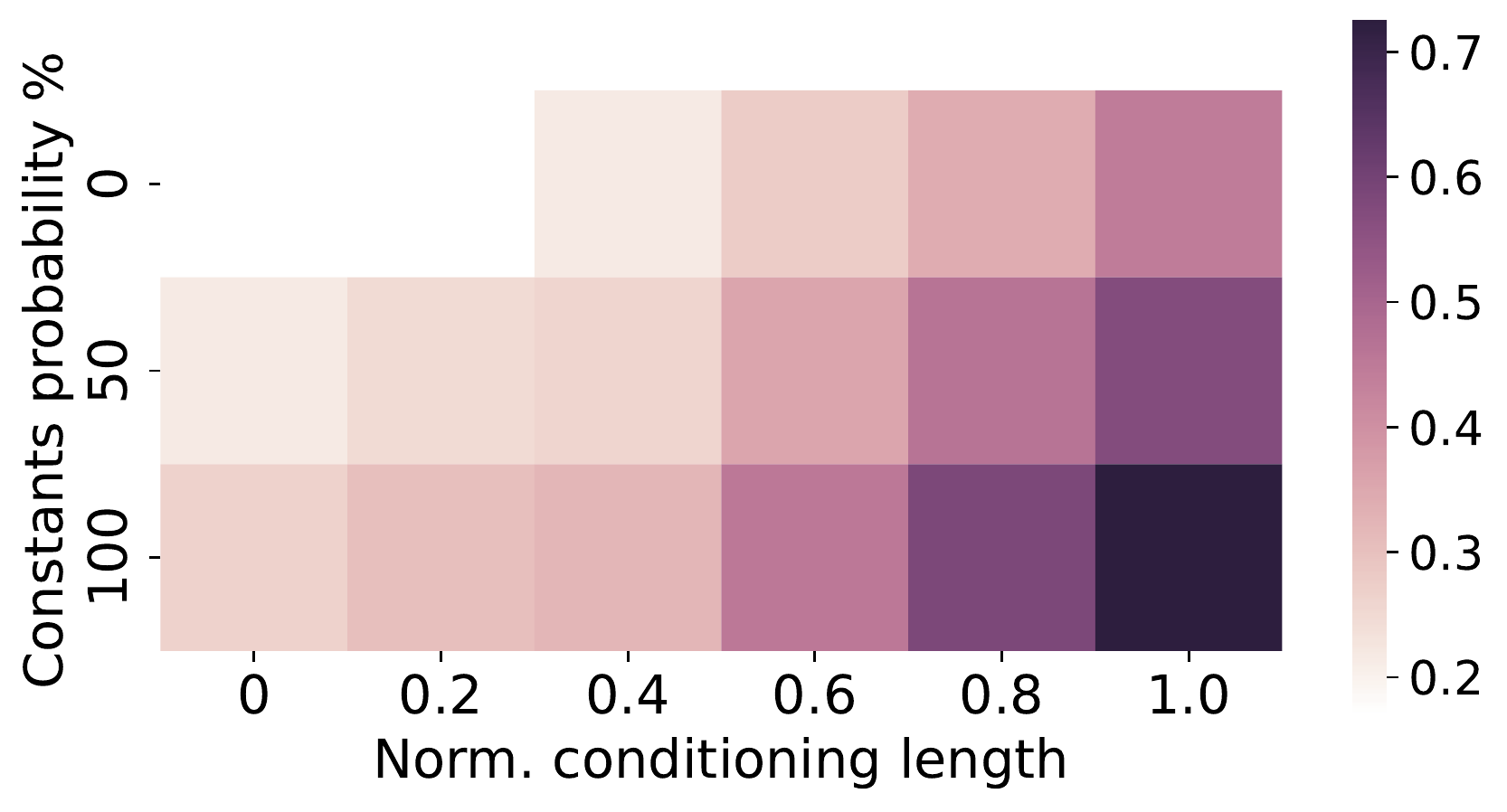}
  \end{center}
\caption{\textbf{Dependence on the amount of conditioning.} \small{The heatmap shows how changing the probability of appearing subtrees and constants affects NSRwH's performance on the \texttt{train$\_$wc} dataset, measured by the \texttt{is$\_$correct} metric. The y-axis shows the probability of constants appearing, with 100\% meaning all constants are inputted. The x-axis shows the normalized conditioning length, with 1.0 meaning the model sees positive sub-branches whose length adds up to the prefix ground truth.}}\label{heatmap}
\end{wrapfigure}
In this section, we investigate how the performance changes as we increase the amount of conditioning. We conduct this experiment using both \texttt{Positive} and \texttt{Constants} as we can easily control the degree of conditioning by adjusting the probability of the number of subtrees and constants that appear, respectively. As before, the beam size for both NSRwH and NeSymReS is set to 5. Fig \ref{heatmap} shows how the value of the \texttt{is$\_$correct} metric changes as we vary the amount of \texttt{Positive} and \texttt{Constants} information. 
As expected, a monotonic trend can be observed for both properties as the amount of conditioning is increased. The peak in performance is reached when the two properties are provided in the largest amounts, suggesting that the model can combine the two prompts to maximize its prediction accuracy.

\subsection{What if no assumptions can be made?}\label{pierre}
This section investigates the scenario where no prior knowledge is available to condition the model. 
\begin{wraptable}{r}{0.5\textwidth}
\resizebox{0.5\textwidth}{!}{%
\begin{tabular}{ccc}
\hline
{Model Type}        & \texttt{is$\_$correct}  & \texttt{$R^2_{\text{mean}}$}     \\ \hline
Random Positive Conditions & \textbf{0.35 $\pm$ 0.06} & \textbf{0.86 $\pm$ 0.05} \\
Standard Model {[}5K{]}    & 0.23 $\pm$ 0.00          & 0.74 $\pm$ 0.05          \\
Standard Model {[}10{]}    & 0.12 $\pm$ 0.04          & 0.32 $\pm$ 0.03           \\ \hline
\end{tabular}
}
\caption{\textbf{No privileged information available.} Comparison between NSRwH with randomly sampled hypotheses, a standard NSR approach (NeSymReS) with beam size 5000, and a standard NSR approach with beam size 10 (same as NSRwH). Results are averaged over 5 runs.}\label{exp3}
\end{wraptable}
The objective of the experiment is to determine if using NSRwH with randomly sampled hypotheses can outperform a standard NSR model, which can only improve its predictions by increasing the beam size. According to prior work, conventional search techniques of NSR, such as beam search and random sampling, quickly reach a saturation point in exploring the search space, making larger beam sizes ineffective for exploration (see Fig. 16 in \citep{kamienny2022end}). The experiment is conducted on the \texttt{black$\_$box} dataset. The standard model uses a large beam size of $M=5000$, which is within the saturation regime, and NSRwH uses $N=500$ diverse, randomly sampled \texttt{Positive} conditionings with a beam size of $M/N=10$ for each. As such, both methods utilize the same computational budget. Table \ref{exp3} shows that NSRwH outperforms the standard NSR model on the \texttt{black$\_$box} dataset. We highlight that the policy used to randomly sample positive operators is very sparse and highly suboptimal. As such, the design of more effective search routines over the space of properties represents an interesting avenue for future research.

\section{Discussion}\label{sec:conc}
\paragraph{Conclusive remarks.}
This work presents a novel approach for symbolic regression that enables the explicit incorporation of inductive biases reflecting prior knowledge of the problem at hand. In contrast to previous works, this can be effectively done at test time, drastically reducing the computational overhead. Thanks to this property, our model better lends itself to online and interactive applications of symbolic regression, thus enabling fast hypothesis testing, a highly desirable feature for scientific discovery applications. We demonstrate the value of this approach with a number of examples and ablation studies where numerical data is scarce or affected by noise. 

\paragraph{Limitations and future work}
The main limitation of the proposed approach is realized in the scenario where no prior knowledge is available. In this case, the performance gains obtained in Section \ref{sec:privabailable} are not guaranteed. However, in Section \ref{pierre}, our final experiment suggests an intriguing opportunity for future research - leveraging NSRwH's extra degree of freedom to explore the equation space more efficiently. In addition, the properties investigated in this work are not exhaustive and it is conceivable to include additional forms of prior knowledge, such as alternative definitions of the complexity of mathematical expressions based on syntax or semantics \citep{complexity,Vladislavleva2009OrderON}. Finally, we remark that thanks to its simplicity, the same idea at the basis of NSRwH can be applied to more advanced NSR algorithms, like the one recently proposed by \cite{kamienny2022end}, likely resulting in further performance improvements. We intend to investigate the above questions in future work. 

\bibliography{main}

\begin{thebibliography}{38}
\providecommand{\natexlab}[1]{#1}
\providecommand{\url}[1]{\texttt{#1}}
\expandafter\ifx\csname urlstyle\endcsname\relax
  \providecommand{\doi}[1]{doi: #1}\else
  \providecommand{\doi}{doi: \begingroup \urlstyle{rm}\Url}\fi

\bibitem[Becker et~al.()Becker, Klein, Neitz, Parascandolo, and
  Kilbertus]{becker2022discovering}
S{\"o}ren Becker, Michal Klein, Alexander Neitz, Giambattista Parascandolo, and
  Niki Kilbertus.
\newblock Discovering ordinary differential equations that govern time-series.
\newblock In \emph{NeurIPS 2022 AI for Science: Progress and Promises}.

\bibitem[Biggio et~al.(2020)Biggio, Bendinelli, Lucchi, and
  Parascandolo]{biggio2020seq2seq}
Luca Biggio, Tommaso Bendinelli, Aurelien Lucchi, and Giambattista
  Parascandolo.
\newblock A seq2seq approach to symbolic regression.
\newblock In \emph{Learning Meets Combinatorial Algorithms at NeurIPS2020},
  2020.

\bibitem[Biggio et~al.(2021)Biggio, Bendinelli, Neitz, Lucchi, and
  Parascandolo]{biggio2021neural}
Luca Biggio, Tommaso Bendinelli, Alexander Neitz, Aurelien Lucchi, and
  Giambattista Parascandolo.
\newblock Neural symbolic regression that scales, 2021.

\bibitem[Brunton et~al.(2016)Brunton, Proctor, and
  Kutz]{brunton2016discovering}
Steven~L Brunton, Joshua~L Proctor, and J~Nathan Kutz.
\newblock Discovering governing equations from data by sparse identification of
  nonlinear dynamical systems.
\newblock \emph{Proceedings of the national academy of sciences}, 113\penalty0
  (15):\penalty0 3932--3937, 2016.

\bibitem[Burlacu et~al.(2020)Burlacu, Kronberger, and
  Kommenda]{10.1145/3377929.3398099}
Bogdan Burlacu, Gabriel Kronberger, and Michael Kommenda.
\newblock Operon c++: An efficient genetic programming framework for symbolic
  regression.
\newblock In \emph{Proceedings of the 2020 Genetic and Evolutionary Computation
  Conference Companion}, GECCO '20, page 1562–1570, New York, NY, USA, 2020.
  Association for Computing Machinery.
\newblock ISBN 9781450371278.
\newblock \doi{10.1145/3377929.3398099}.
\newblock URL \url{https://doi.org/10.1145/3377929.3398099}.

\bibitem[Charton(2022)]{charton2022my}
Fran{\c{c}}ois Charton.
\newblock What is my math transformer doing?--three results on interpretability
  and generalization.
\newblock \emph{arXiv preprint arXiv:2211.00170}, 2022.

\bibitem[Cranmer et~al.(2020)Cranmer, Sanchez~Gonzalez, Battaglia, Xu, Cranmer,
  Spergel, and Ho]{cranmer2020discovering}
Miles Cranmer, Alvaro Sanchez~Gonzalez, Peter Battaglia, Rui Xu, Kyle Cranmer,
  David Spergel, and Shirley Ho.
\newblock Discovering symbolic models from deep learning with inductive biases.
\newblock \emph{Advances in Neural Information Processing Systems},
  33:\penalty0 17429--17442, 2020.

\bibitem[d'Ascoli et~al.(2022)d'Ascoli, Kamienny, Lample, and
  Charton]{d2022deep}
St{\'e}phane d'Ascoli, Pierre-Alexandre Kamienny, Guillaume Lample, and
  Fran{\c{c}}ois Charton.
\newblock Deep symbolic regression for recurrent sequences.
\newblock \emph{arXiv preprint arXiv:2201.04600}, 2022.

\bibitem[Fletcher(1987)]{Flet87}
Roger Fletcher.
\newblock \emph{Practical Methods of Optimization}.
\newblock John Wiley \& Sons, New York, NY, USA, second edition, 1987.

\bibitem[Haider et~al.(2022)Haider, de~Fran{\c{c}}a, Kronberger, and
  Burlacu]{haider2022comparing}
Christian Haider, Fabr{\'\i}cio~Olivetti de~Fran{\c{c}}a, Gabriel Kronberger,
  and Bogdan Burlacu.
\newblock Comparing optimistic and pessimistic constraint evaluation in
  shape-constrained symbolic regression.
\newblock In \emph{Proceedings of the Genetic and Evolutionary Computation
  Conference}, pages 938--945, 2022.

\bibitem[Hernandez et~al.(2019)Hernandez, Balasubramanian, Yuan, Mason, and
  Mueller]{hernandez2019fast}
Alberto Hernandez, Adarsh Balasubramanian, Fenglin Yuan, Simon~AM Mason, and
  Tim Mueller.
\newblock Fast, accurate, and transferable many-body interatomic potentials by
  symbolic regression.
\newblock \emph{npj Computational Materials}, 5\penalty0 (1):\penalty0 1--11,
  2019.

\bibitem[Kabliman et~al.(2021)Kabliman, Kolody, Kronsteiner, Kommenda, and
  Kronberger]{kabliman2021application}
Evgeniya Kabliman, Ana~Helena Kolody, Johannes Kronsteiner, Michael Kommenda,
  and Gabriel Kronberger.
\newblock Application of symbolic regression for constitutive modeling of
  plastic deformation.
\newblock \emph{Applications in Engineering Science}, 6:\penalty0 100052, 2021.

\bibitem[Kaheman et~al.(2020)Kaheman, Kutz, and Brunton]{kaheman2020sindy}
Kadierdan Kaheman, J~Nathan Kutz, and Steven~L Brunton.
\newblock Sindy-pi: a robust algorithm for parallel implicit sparse
  identification of nonlinear dynamics.
\newblock \emph{Proceedings of the Royal Society A}, 476\penalty0
  (2242):\penalty0 20200279, 2020.

\bibitem[Kamienny et~al.(2022)Kamienny, d'Ascoli, Lample, and
  Charton]{kamienny2022end}
Pierre-Alexandre Kamienny, St{\'e}phane d'Ascoli, Guillaume Lample, and
  Fran{\c{c}}ois Charton.
\newblock End-to-end symbolic regression with transformers.
\newblock \emph{arXiv preprint arXiv:2204.10532}, 2022.

\bibitem[Kommenda et~al.(2015)Kommenda, Beham, Affenzeller, and
  Kronberger]{complexity}
Michael Kommenda, Andreas Beham, Michael Affenzeller, and Gabriel Kronberger.
\newblock Complexity measures for multi-objective symbolic regression.
\newblock In \emph{Computer Aided Systems Theory {\textendash} {EUROCAST}
  2015}, pages 409--416. Springer International Publishing, 2015.
\newblock \doi{10.1007/978-3-319-27340-2_51}.
\newblock URL \url{https://doi.org/10.1007%2F978-3-319-27340-2_51}.

\bibitem[Kronberger et~al.(2022)Kronberger, de~Fran{\c{c}}a, Burlacu, Haider,
  and Kommenda]{kronberger2022shape}
Gabriel Kronberger, Fabricio~Olivetti de~Fran{\c{c}}a, Bogdan Burlacu,
  Christian Haider, and Michael Kommenda.
\newblock Shape-constrained symbolic regression—improving extrapolation with
  prior knowledge.
\newblock \emph{Evolutionary Computation}, 30\penalty0 (1):\penalty0 75--98,
  2022.

\bibitem[La~Cava et~al.(2021)La~Cava, Orzechowski, Burlacu, de~França,
  Virgolin, Jin, Kommenda, and Moore]{lacava}
William La~Cava, Patryk Orzechowski, Bogdan Burlacu, Fabrício~Olivetti
  de~França, Marco Virgolin, Ying Jin, Michael Kommenda, and Jason~H. Moore.
\newblock Contemporary symbolic regression methods and their relative
  performance, 2021.
\newblock URL \url{https://arxiv.org/abs/2107.14351}.

\bibitem[Lample and Charton(2019)]{lample2019deep}
Guillaume Lample and Fran{\c{c}}ois Charton.
\newblock Deep learning for symbolic mathematics.
\newblock \emph{arXiv preprint arXiv:1912.01412}, 2019.

\bibitem[Lee et~al.(2019)Lee, Lee, Kim, Kosiorek, Choi, and Teh]{lee2019set}
Juho Lee, Yoonho Lee, Jungtaek Kim, Adam Kosiorek, Seungjin Choi, and Yee~Whye
  Teh.
\newblock Set transformer: A framework for attention-based
  permutation-invariant neural networks.
\newblock In \emph{International conference on machine learning}, pages
  3744--3753. PMLR, 2019.

\bibitem[Li et~al.(2022)Li, Yuan, and Shen]{Li2022SymbolicET}
Jiachen Li, Ye~Yuan, and Hongze Shen.
\newblock Symbolic expression transformer: A computer vision approach for
  symbolic regression.
\newblock \emph{ArXiv}, abs/2205.11798, 2022.

\bibitem[Ma et~al.(2022)Ma, Narayanaswamy, Riley, and Li]{ma2022evolving}
He~Ma, Arunachalam Narayanaswamy, Patrick Riley, and Li~Li.
\newblock Evolving symbolic density functionals.
\newblock \emph{arXiv preprint arXiv:2203.02540}, 2022.

\bibitem[Meurer et~al.(2017)Meurer, Smith, Paprocki, \v{C}ert\'{i}k, Kirpichev,
  Rocklin, Kumar, Ivanov, Moore, Singh, Rathnayake, Vig, Granger, Muller,
  Bonazzi, Gupta, Vats, Johansson, Pedregosa, Curry, Terrel, Rou\v{c}ka, Saboo,
  Fernando, Kulal, Cimrman, and Scopatz]{sympy}
Aaron Meurer, Christopher~P. Smith, Mateusz Paprocki, Ond\v{r}ej
  \v{C}ert\'{i}k, Sergey~B. Kirpichev, Matthew Rocklin, AMiT Kumar, Sergiu
  Ivanov, Jason~K. Moore, Sartaj Singh, Thilina Rathnayake, Sean Vig, Brian~E.
  Granger, Richard~P. Muller, Francesco Bonazzi, Harsh Gupta, Shivam Vats,
  Fredrik Johansson, Fabian Pedregosa, Matthew~J. Curry, Andy~R. Terrel,
  \v{S}t\v{e}p\'{a}n Rou\v{c}ka, Ashutosh Saboo, Isuru Fernando, Sumith Kulal,
  Robert Cimrman, and Anthony Scopatz.
\newblock Sympy: symbolic computing in python.
\newblock \emph{PeerJ Computer Science}, 3:\penalty0 e103, January 2017.
\newblock ISSN 2376-5992.
\newblock \doi{10.7717/peerj-cs.103}.
\newblock URL \url{https://doi.org/10.7717/peerj-cs.103}.

\bibitem[Mundhenk et~al.(2021)Mundhenk, Landajuela, Glatt, Santiago, Faissol,
  and Petersen]{mundhenk2021symbolic}
T~Nathan Mundhenk, Mikel Landajuela, Ruben Glatt, Claudio~P Santiago, Daniel~M
  Faissol, and Brenden~K Petersen.
\newblock Symbolic regression via neural-guided genetic programming population
  seeding.
\newblock \emph{arXiv preprint arXiv:2111.00053}, 2021.

\bibitem[Olson et~al.(2017)Olson, La~Cava, Orzechowski, Urbanowicz, and
  Moore]{Olson2017PMLB}
Randal~S. Olson, William La~Cava, Patryk Orzechowski, Ryan~J. Urbanowicz, and
  Jason~H. Moore.
\newblock Pmlb: a large benchmark suite for machine learning evaluation and
  comparison.
\newblock \emph{BioData Mining}, 10\penalty0 (1):\penalty0 36, Dec 2017.
\newblock ISSN 1756-0381.
\newblock \doi{10.1186/s13040-017-0154-4}.
\newblock URL \url{https://doi.org/10.1186/s13040-017-0154-4}.

\bibitem[Ramesh et~al.(2022)Ramesh, Dhariwal, Nichol, Chu, and
  Chen]{ramesh2022hierarchical}
Aditya Ramesh, Prafulla Dhariwal, Alex Nichol, Casey Chu, and Mark Chen.
\newblock Hierarchical text-conditional image generation with clip latents.
\newblock \emph{arXiv preprint arXiv:2204.06125}, 2022.

\bibitem[Saharia et~al.(2022)Saharia, Chan, Saxena, Li, Whang, Denton,
  Ghasemipour, Ayan, Mahdavi, Lopes, et~al.]{saharia2022photorealistic}
Chitwan Saharia, William Chan, Saurabh Saxena, Lala Li, Jay Whang, Emily
  Denton, Seyed Kamyar~Seyed Ghasemipour, Burcu~Karagol Ayan, S~Sara Mahdavi,
  Rapha~Gontijo Lopes, et~al.
\newblock Photorealistic text-to-image diffusion models with deep language
  understanding.
\newblock \emph{arXiv preprint arXiv:2205.11487}, 2022.

\bibitem[Schmidt and Lipson(2009)]{schmidt2009distilling}
Michael Schmidt and Hod Lipson.
\newblock Distilling free-form natural laws from experimental data.
\newblock \emph{science}, 324\penalty0 (5923):\penalty0 81--85, 2009.

\bibitem[Strogatz(2018)]{strogatz2018nonlinear}
Steven~H Strogatz.
\newblock \emph{Nonlinear dynamics and chaos: with applications to physics,
  biology, chemistry, and engineering}.
\newblock CRC press, 2018.

\bibitem[Sun et~al.(2022)Sun, Liu, Wang, and Sun]{sun2022symbolic}
Fangzheng Sun, Yang Liu, Jian-Xun Wang, and Hao Sun.
\newblock Symbolic physics learner: Discovering governing equations via monte
  carlo tree search.
\newblock \emph{arXiv preprint arXiv:2205.13134}, 2022.

\bibitem[Sutskever et~al.(2014)Sutskever, Vinyals, and
  Le]{sutskever2014sequence}
Ilya Sutskever, Oriol Vinyals, and Quoc~V Le.
\newblock Sequence to sequence learning with neural networks.
\newblock \emph{Advances in neural information processing systems}, 27, 2014.

\bibitem[Udrescu and Tegmark(2020)]{udrescu2020ai}
Silviu-Marian Udrescu and Max Tegmark.
\newblock Ai feynman: A physics-inspired method for symbolic regression.
\newblock \emph{Science Advances}, 6\penalty0 (16):\penalty0 eaay2631, 2020.

\bibitem[Udrescu et~al.(2020)Udrescu, Tan, Feng, Neto, Wu, and
  Tegmark]{udrescu2}
Silviu-Marian Udrescu, Andrew Tan, Jiahai Feng, Orisvaldo Neto, Tailin Wu, and
  Max Tegmark.
\newblock Ai feynman 2.0: Pareto-optimal symbolic regression exploiting graph
  modularity.
\newblock 2020.
\newblock \doi{10.48550/ARXIV.2006.10782}.
\newblock URL \url{https://arxiv.org/abs/2006.10782}.

\bibitem[Vaddireddy et~al.(2020)Vaddireddy, Rasheed, Staples, and
  San]{vaddireddy2020feature}
Harsha Vaddireddy, Adil Rasheed, Anne~E Staples, and Omer San.
\newblock Feature engineering and symbolic regression methods for detecting
  hidden physics from sparse sensor observation data.
\newblock \emph{Physics of Fluids}, 32\penalty0 (1):\penalty0 015113, 2020.

\bibitem[Valipour et~al.(2021)Valipour, You, Panju, and
  Ghodsi]{valipour2021symbolicgpt}
Mojtaba Valipour, Bowen You, Maysum Panju, and Ali Ghodsi.
\newblock Symbolicgpt: A generative transformer model for symbolic regression.
\newblock \emph{arXiv preprint arXiv:2106.14131}, 2021.

\bibitem[Vastl et~al.(2022)Vastl, Kulh{\'a}nek, Kubal{\'i}k, Derner, and
  Babu{$\vs$}ka]{Vastl2022SymFormerES}
Martin Vastl, Jon{\'a}s Kulh{\'a}nek, Jir{\'i} Kubal{\'i}k, Erik Derner, and
  Robert Babu{$\vs$}ka.
\newblock Symformer: End-to-end symbolic regression using transformer-based
  architecture.
\newblock \emph{ArXiv}, abs/2205.15764, 2022.

\bibitem[Vaswani et~al.(2017)Vaswani, Shazeer, Parmar, Uszkoreit, Jones, Gomez,
  Kaiser, and Polosukhin]{vaswani2017attention}
Ashish Vaswani, Noam Shazeer, Niki Parmar, Jakob Uszkoreit, Llion Jones,
  Aidan~N Gomez, {\L}ukasz Kaiser, and Illia Polosukhin.
\newblock Attention is all you need.
\newblock \emph{Advances in neural information processing systems}, 30, 2017.

\bibitem[Vladislavleva et~al.(2009)Vladislavleva, Smits, and den
  Hertog]{Vladislavleva2009OrderON}
Ekaterina Vladislavleva, Guido Smits, and Dick den Hertog.
\newblock Order of nonlinearity as a complexity measure for models generated by
  symbolic regression via pareto genetic programming.
\newblock \emph{IEEE Transactions on Evolutionary Computation}, 13:\penalty0
  333--349, 2009.

\bibitem[Wang et~al.(2019)Wang, Wagner, and Rondinelli]{wang2019symbolic}
Yiqun Wang, Nicholas Wagner, and James~M Rondinelli.
\newblock Symbolic regression in materials science.
\newblock \emph{MRS Communications}, 9\penalty0 (3):\penalty0 793--805, 2019.

\end{thebibliography}
\bibliographystyle{plainnat}

%%%%%%%%%%%%%%%%%%%%%%%%%%%%%%%%%%%%%%%%%%%%%%%%%%%%%%%%%%%%%%%%%%%%%%%%%%%%%%%
%%%%%%%%%%%%%%%%%%%%%%%%%%%%%%%%%%%%%%%%%%%%%%%%%%%%%%%%%%%%%%%%%%%%%%%%%%%%%%%
% APPENDIX
%%%%%%%%%%%%%%%%%%%%%%%%%%%%%%%%%%%%%%%%%%%%%%%%%%%%%%%%%%%%%%%%%%%%%%%%%%%%%%%
%%%%%%%%%%%%%%%%%%%%%%%%%%%%%%%%%%%%%%%%%%%%%%%%%%%%%%%%%%%%%%%%%%%%%%%%%%%%%%%
\newpage
\appendix
\onecolumn
\section{Dataset generation}\label{app1}

\subsection{Generating $\mathcal{D}$}\label{app1_1}
We build our training dataset first by generating symbolic expressions skeletons (i.e. mathematical expressions where the values of the constants are replaced by placeholder tokens) using the framework introduced by \cite{lample2019deep}. Our vocabulary consists of the unary and binary operators shown in Table \ref{tab:operators}. We considered scalar (output dimension equal to 1) expressions with up to 5 independent variables with a maximum prefix length and depth of 20 and 6 respectively.

To obtain the mathematical expression and corresponding numerical evaluation during training for each equation we adopt the following procedure: 

\begin{itemize}
    \item  An equation skeleton is randomly sampled from the pool of symbolic expression skeletons
    \item The sampled expression is simplified using the \texttt{simplify} function from \texttt{Sympy} \cite{sympy} in order to remove any redundant term.
    \item Constants of the skeleton are sampled from $\mathcal{U}(-10,-10)$ if they are additive, and logarithmically from $\mathcal{U}(0.05,10)$ if multiplicative. 

    \item The extrema of the support for each independent variable is sampled independently from a uniform distribution $\mathcal{U}(-10,10)$ with the distance between the left and right extrema of at least $1$.

    \item For each independent variable, $n$ input points are sampled from the previously sampled support, where $n$ is sampled between $\mathcal{U}(1,1000)$.
    Support points that lead to absolute values bigger than $65504$ or NANs are discarded and re-sampled. 

    \item We evaluate the sampled expression on the previously obtained support points by using the \texttt{lambdify} function from \texttt{Sympy} \citep{sympy}.
\end{itemize}

\begin{table}[H]
    \centering
    \small
    \begin{tabular}{c|c}
    \toprule
        Arity & Operators \\
        \midrule
        Unary & 
        \texttt{sqrt, pow2, pow3, pow4}\\
        & \texttt{inv, log, exp}\\
        & \texttt{sin, cos, asin} \\
        \hline
        Binary & \texttt{add, sub, mul, div}\\
    \bottomrule
    \end{tabular}
    \caption{Operators used in our data generation pipeline.}
    \label{tab:operators}
\end{table}

As the input evaluations can lead to large values, we follow \cite{biggio2021neural} and we convert them from float to a multi-hot bit representation according to the half-precision IEEE-754 standard before feeding them into the model.

\subsection{Generating  $\privileged$}\label{app1_2}

\subsubsection{Complexity}
The complexity of a sentence is determined by the sum of the number of nodes and leaves in the expression, as outlined in \cite{lacava}. Each complexity value is represented by a unique token, ranging from 1 (i.e. $x_1$) to 20. 

\subsubsection{Symmetry}
We use the definition of generalized symmetry proposed by \cite{udrescu2}: $f$ has generalized symmetry if the $d$ components of the vector $\mathbf{x} \in \mathbb{R}^d$ can be split into groups of $k$ and $d-k$ components (which we denote by the vectors $\mathbf{x}^{\prime} \in \mathbb{R}^k$ and $\mathbf{x}^{\prime \prime} \in \mathbb{R}^{d-k}$ ) such that $f(\mathbf{x})=f\left(\mathbf{x}^{\prime}, \mathbf{x}^{\prime \prime}\right)=g\left[h\left(\mathbf{x}^{\prime}\right), \mathbf{x}^{\prime \prime}\right]$ for some unknown function $g$. 
As explained in \cite{udrescu2}, in order to check the presence of generalized symmetry in the set of variables $\mathbf{x'}$, it is sufficient to check whether the normalized gradient of $f$ with respect to $\mathbf{x'}$ is independent on $\mathbf{x''}$, i.e.  $\frac{{\nabla_{\mathbf{x}^{\prime}}} f\left(\mathbf{x}^{\prime}, \mathbf{x}^{\prime \prime}\right)}{\vert{\nabla_{\mathbf{x}^{\prime}}} f\left(\mathbf{x}^{\prime}, \mathbf{x}^{\prime \prime}\right)\vert}$ is $\mathbf{x''}$-independent.
We have created two tokens for each symmetry combination, one to represent the presence of symmetry and one to represent its absence. The total number of tokens is 50, as there are 32 possible symmetry combinations when there are five variables, but some of them are not informative and are excluded, leaving 25 useful combinations. When the number of variables is less than five, only the tokens related to the actual variables are passed to the model (see example in Section \ref{app1_exa}).

\subsubsection{Appearing / absent branches}\label{appearing}
To sample both appearing and absent branches for an expression, we create two candidate pools: a positive and a negative one. The positive pool is created by using the Depth-First-Search (DFS) algorithm to list all the subtrees within the current prefix expression and then by removing the subtree corresponding to the entire expression and other non-informative subtrees like $x_i$. The negative pool is created by filtering out the branches that are already present in the current prefix tree from a pre-computed set of branches, obtained from a large pool of expression trees within the training distribution. 
We sample subtrees from these pools with a probability proportional to the inverse of their length squared, both during training and evaluation. To regulate the total information given to the model, two parameters, $p_p$ for the positive subtrees and $p_n$ for the negative subtrees, are used. The product of $p_p$ ($p_n$) and the ground truth length determines the total number of tokens provided to the model, denoted as $s_p$ ($s_n$). Positive (Negative) subtrees are sampled until the aggregate sum of their lengths, $\sum_{i=1}^{N} \texttt{len}(\text{sampled subtrees}_i)$, reaches the $s_p$ ($s_n$) value. Sub-branches are separated by special separator tokens.

\subsubsection{Constants}\label{constants}
Each a-priori-known constant is assigned to a specific symbol, such as \texttt{pointer$\_$0} for the first constant, \texttt{pointer$\_$1} for the second, and so on. We then give the symbolic encoder the corresponding pointer and a numerical embedding obtained by first converting the known constant in its equivalent 16-bit representation and then passing it through a learnable linear layer that makes its dimension match that of the symbolic embedding. In the target expression, we replace the standard constant placeholder with the \texttt{pointer$\_i$} token in the expression. At training and evaluation stages, we regulate the probability of a constant being a-priori-known with a parameter $p_{c}$.

\subsubsection{Example of extraction and processing of conditioning information}\label{app1_exa}
This section provides a concrete example of how different conditionings are extracted and processed to be fed into our model. Consider the expression $x_3 \sin \left(x_1+x_2\right)$.\\
To determine the \texttt{Positive} conditioning, we must first convert it into prefix notation. This is achieved by first rewriting it as ['mul', '$x_3$', 'sin', 'add', '$x_1$', '$x_2$'] and then enumerating all the possible subtrees of the expression, excluding trivial subtrees such as '$x_2$' alone. These are: [['add'], ['mul'], ['sin'], ['add', '$x_1$'], ['add', '$x_2$'], ['mul', '$x_3$'], ['add', '$x_1$', '$x_2$'], ['sin', 'add', '$x_1$', '$x_2$'], ['mul', 'sin', 'add', '$x_1$', '$x_2$']]. Positive conditionings are then sampled from this pool with a probability inversely proportional to the length squared of the subtree. So a positive conditioning such as ['mul', '$x_3$'] is less likely to be sampled than ['sin'] but more likely than ['mul', 'sin', 'add', '$x_1$', '$x_2$'].\\
To obtain the \texttt{Negative} conditionings, we generate subtrees at random that are absent from the positive pool. This is achieved by randomly selecting an expression, enumerating the subtrees within it, and then randomly choosing subtrees from the expression that are not present in the positive pool. For example, for the expression above, a negative conditioning could be ['mul', '$x_1$'], or ['exp'] since none of these are present in the positive pool. The number of sub-trees supplied to the model is determined by the values of $p_p$ and $p_n$, and the total length of the expression. For example, if $p_p$ is 0.5, then the total length of the sampled sub-trees will be 3, since the overall length of the ground truth is 6.\\
\texttt{Constant} conditioning would be empty since no constants can be obtained.\\
\texttt{Complexity} conditioning is simply the sum of total nodes and leaves of the prefix expression tree, so in this case, it is equal to 6.\\
For the \texttt{Symmetry} conditioning, we followed the definition given provided by \cite{udrescu2}. For our example expression, we will have symmetry between $x_1$ and $x_2$ but not between $x_1,x_3$ or $x_2,x_3$.\\
Once computed, the conditionings are wrapped into a string, tokenized, and then fed into the model. The string will have the following form:

[<Positive>, 'sin', </Positive>, <Positive>, 'mul', '$x_3$', </Positive>, <Negative>, 'exp', </Negative>, <Negative>, 'mul', '$x_1$', </Negative>, 'Complexity=6', 'TrueSymmetryX1X2', 'FalseSymmetryX2X3', 'FalseSymmetryX1X3'].\\

\vspace{-0.3cm}

If some conditionings should be masked, they are simply excluded from the list; for instance, if we only want to provide symmetry conditioning, the string would have the following form:

['TrueSymmetryX1X2', 'FalseSymmetryX2X3', 'FalseSymmetryX1X3']

\section{Training and testing details}\label{app3}
We trained the model with 200 million equations using three NVIDIA GeForce RTX 3090 for a total of five days with a batch size of 400. 
As in \cite{biggio2021neural}, we used a 5-layer set encoder as our numerical encoder and a five-layer standard Transformer decoder as our expression generator. The conditioning and numerical embedding are summed before the expression generator.\\
In the training process, the Adam optimizer is employed to optimize the cross-entropy loss, utilizing an initial learning rate of $10^{-4}$, which is subsequently adjusted in proportion to the inverse square root of the number of steps taken.\\
To ensure a fair comparison, the standard model, \texttt{standard$\_$nesy}, was trained using the same number of equations and the same numerical encoder and expression generator architecture. In addition, both models have been trained for an equal number of iterations.

\subsection{Amount of conditioning during training}
During training, we give the model a varying amount of conditioning signals to avoid excessive dependence on them. We adopt the following approach:

\begin{itemize}
    \item \texttt{Positive}: $p_p$, as defined in the sub-section \ref{appearing}, is 0 with probability 0.7. Otherwise, it is sampled from $\mathcal{U}(0,1)$ 

    \item \texttt{Negative}: $p_n$, as defined in the sub-section \ref{appearing} is 0 with probability 0.7. Otherwise, it is sampled from $\mathcal{U}(0,1)$

    \item \texttt{Complexity}: We provide the complexity token to the network with a probability of $0.3$

  \item \texttt{Symmetry}: We provide the symmetry tokens to the network with a probability of $0.2$.

  \item \texttt{Constants}: $p_{c}$ as defined in the sub-section \ref{constants} is equal to $0.15$. 
\end{itemize}

\subsection{Amount of conditioning during testing}
We use a variety of conditioning signals, with each combination of signals referred to by a specific term.

\begin{itemize}
    \item \texttt{Positive}:  $p_p$ as defined in the sub-section \ref{appearing} is equal to 0.5. The other conditioning signals are disabled. 

    \item \texttt{Negative}: $p_n$ as defined in the sub-section \ref{appearing} is equal to 0.5. The other conditioning signals are disabled. 

    \item \texttt{Complexity}: We provide the complexity token to the network. The other conditioning signals are disabled. 

    \item \texttt{Symmetry}: We provide the symmetry token to the network. The other conditioning signals are disabled.
    
    \item \texttt{Constants}: We provide the value of each constant with a probability of $0.8$. The other conditioning signals are disabled. 

    \item \texttt{Vanilla}: No conditioning is given (all conditionings are masked). The model sees only the numerical inputs. This is equivalent to the standard model.

    \item \texttt{All}: combines \texttt{Positive}, \texttt{Negative}, \texttt{Complexity},  \texttt{Symmetry} and \texttt{Constants} conditioning. Each conditioning signal is enabled, with parameters equal to the values mentioned for each individual setting with the sole exception of constants where the probability of providing a constant is set to $0.3$. 
\end{itemize}

\section{Test datasets and metrics}\label{app2}

\subsection{Evaluation datasets}
We created three datasets, \texttt{train$\_$nc}, \texttt{train$\_$wc} and \texttt{only$\_$five$\_$variables$\_$nc} using the same generator configuration as the training set, but with different initial seeds. For \texttt{train$\_$nc} and \texttt{train$\_$wc} datasets, we selected 300 equations, removing all constants from the first and selecting random constants for the second. These equations have different levels of complexity. In contrast, for \texttt{only$\_$five$\_$variables$\_$nc}, we restricted our dataset to equations with five variables, discarding the others. We also removed any constants from these equations. In addition, we evaluate our model on two open-source datasets, namely \texttt{AIF}, consisting of the equation in the AI Feynman database \citep{udrescu2020ai} and  \texttt{black$\_$box} comprising of 14 datasets from the ODE-Strogatz database \citep{strogatz2018nonlinear}. For all experiments except \ref{pierre}, the training points were used to both fit constants
with BFGS and to select the predicted expression among the beam candidates. Specifically,  once the
constants were fitted,  the expression with the lowest BFGS loss was chosen as the predicted expression.
However, since in Section \ref{pierre} a much larger beam size (5000 compared to 5 of the other experiments) was used, we followed a different approach: 60\% of the points were used for fitting constants, and the remaining 40\% to select the best expression. The expression with the highest $R^2$ scores on this 40\% support was chosen as the predicted expression.

\vspace{-0.3cm}
\subsection{Evaluation Procedure}
For \texttt{train$\_$nc}, \texttt{train$\_$wc}  and \texttt{only$\_$five$\_$variables$\_$nc} we sample the support points following the same procedure as in the training pipeline. For AI Feynman equations, we use the support defined in the dataset. For the ODE-Strogatz dataset we followed the approach from \cite{lacava} and used 75\% of the points from the function call \texttt{fetch$\_$data} from the PMLB repository  for training \cite{Olson2017PMLB} and the remaining for testing.\\
For the other datasets, to test the quality of our prediction, we sampled 500 points from the OOD support $\mathcal{U}(-25, 25)$. Our criterion for identifying equations as symbolically equivalent to the ground truth was a 99\% or higher average output of the \texttt{numpy.is\_close($y$,$\hat{y}$)} function across the support points. This threshold accounted for numerical inaccuracies, such as those caused by numerical instability near support points close to zero, so that equations symbolically equivalent were not misclassified due to these errors. 

\vspace{-0.3cm}

\section{Additional results}\label{app4}
\vspace{-0.2cm}
In this section, we report some additional results obtained by evaluating the model on the \texttt{$R^2$} metric. We conclude with a subsection comparing the model obtained by completely masking the symbolic encoder of NSRwH (\texttt{vanilla} model) and a standard NSR model without any symbolic encoder (\texttt{standard$\_$nesy}).

\vspace{-0.3cm}

\subsection{\texttt{$R^2$} metric}
\begin{figure}[H]
\centering
\includegraphics[scale = 0.3]{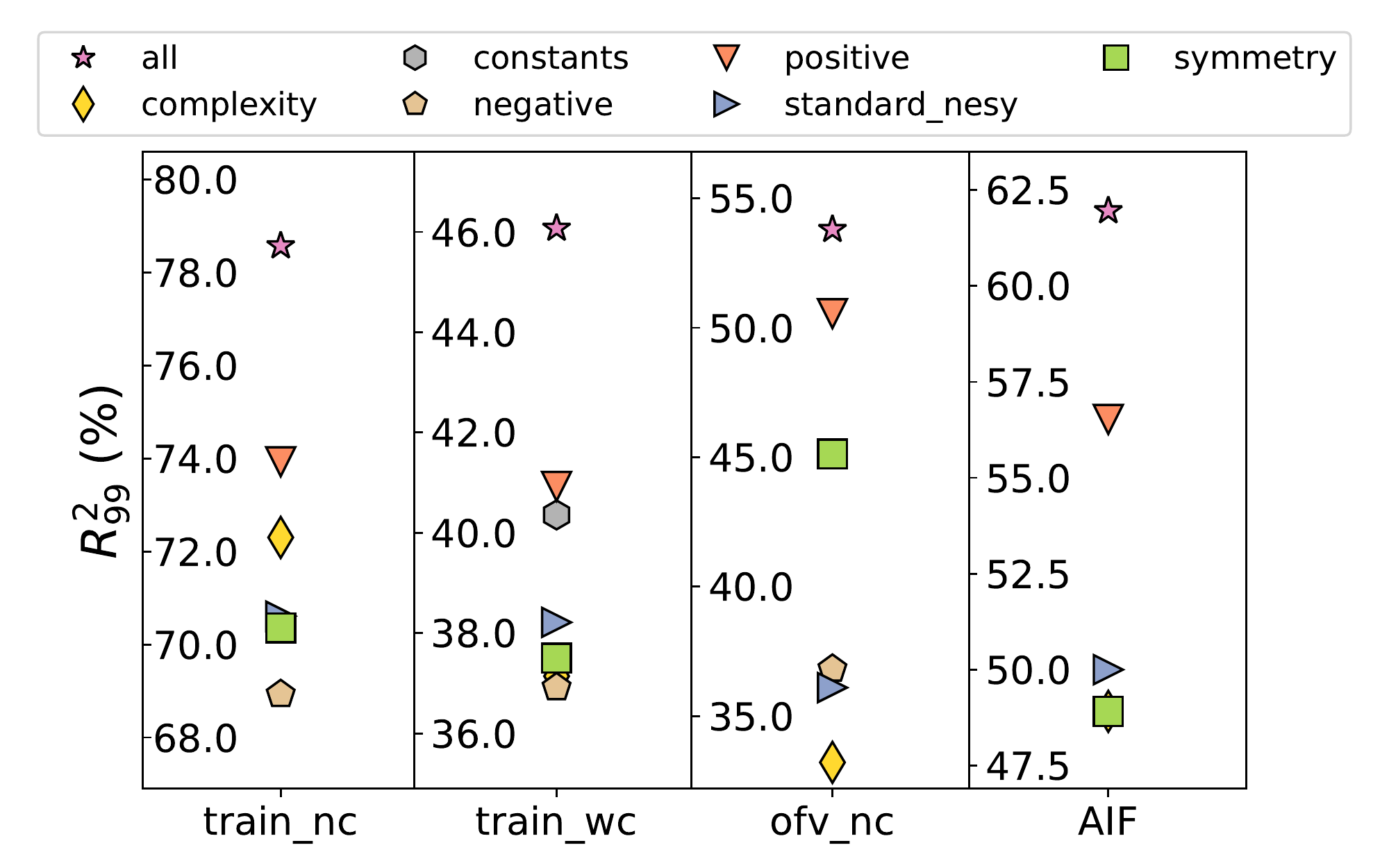}
\caption{\textbf{Conditioning improves performance.} Comparison between NSRwH conditioned with different types of hypotheses and the standard NeSymReS for the different datasets in terms of $R^2_{0.99}$ score} \label{noiseless1}
\end{figure}

\begin{figure}[ht]
\makebox[\textwidth][c]{
\begin{tabular}{cccc}
\includegraphics[scale = 0.24]{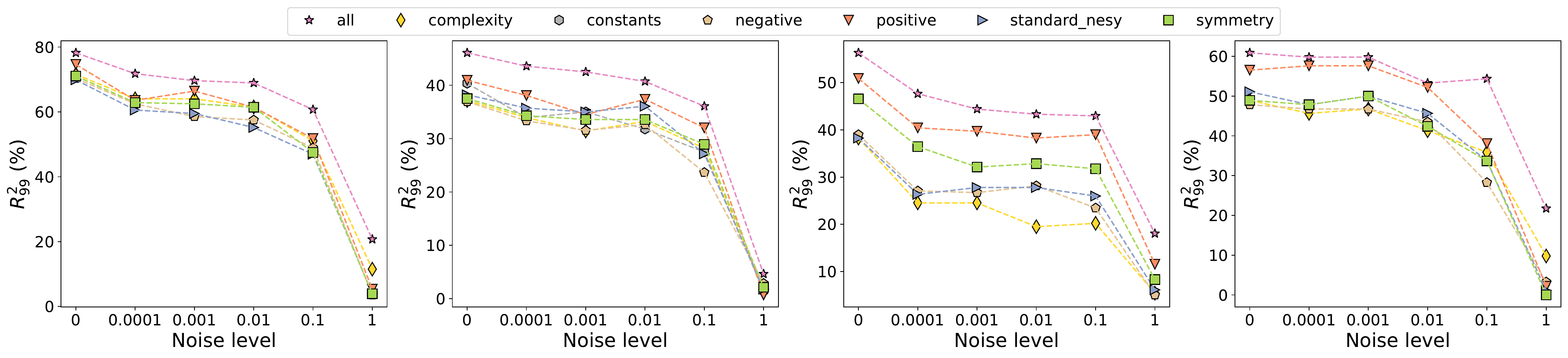}\\
\end{tabular}}
\caption{\textbf{Dependence on the input noise.} Comparison between NSRwH conditioned with different types of hypotheses and the unconditioned baseline (\texttt{standard$\_$nesy}) in terms of the $R^2_{0.99}$ metric, as a function of the noise level in the input data, for the \texttt{train$\_$nc},
\texttt{train$\_$wc}, \texttt{only$\_$five$\_$variables$\_$nc} and \texttt{AIF} datasets from left to right.}\label{noise2}
\end{figure}

\begin{figure}[ht]
\makebox[\textwidth][c]{
\begin{tabular}{cccc}
\includegraphics[scale = 0.24]{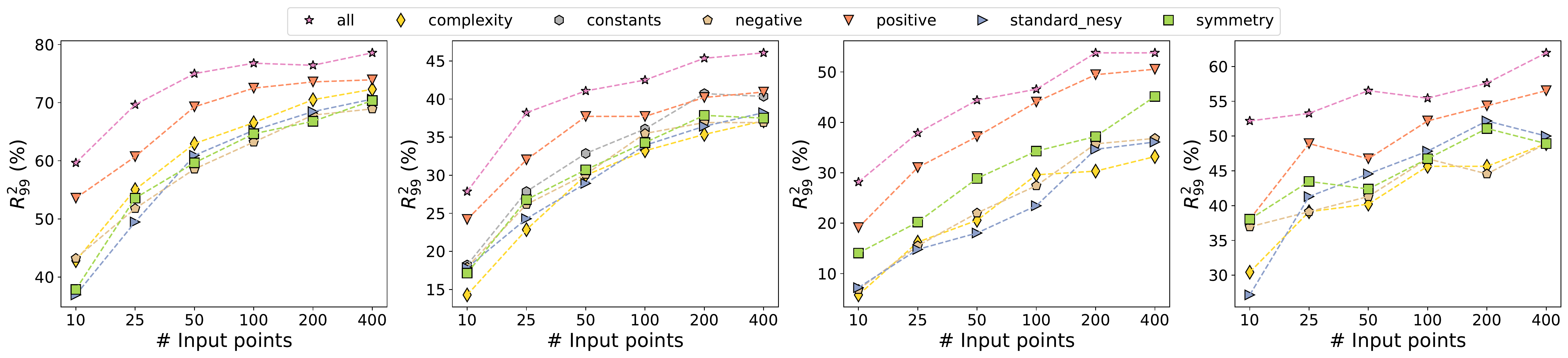}\\
\end{tabular}}
\caption{\textbf{Dependence on the number of input points.} Comparison between NSRwH conditioned with different types of hypotheses and the unconditioned baseline (\texttt{standard$\_$nesy}) in terms of the $R^2_{0.99}$ metric, as a function of the number of input points, for the \texttt{train$\_$nc},
\texttt{train$\_$wc}, \texttt{only$\_$five$\_$variables$\_$nc} and \texttt{AIF} datasets from left to right.}\label{points2}
\end{figure}

\vspace{-0.2cm}
Figs \ref{noiseless1}, \ref{noise2} and \ref{points2} repeat the analysis performed in the main body but with the \texttt{$R^2$} metric instead of the \texttt{is$\_$correct} metric. The scores in this section are calculated by extracting the $R^2$ value for each expression. If such a value is above 0.99, a score of 1 is assigned, otherwise zero. Finally, the so obtained boolean scores are averaged across the entire test set. We refer to this metric as $R^2_{0.99}$.

\vspace{-0.3cm}

\vspace{-0.5cm}
\subsection{Comparison between masked NSRwH and standard model}
\vspace{-0.3cm}
In this section, we compare the fully masked model -- referred to as \texttt{vanilla} -- to the standard NSR method (without a symbolic encoder) -- referred to as \texttt{standard$\_$nesy}. The goal is to show that their performance is aligned, indicating that NSRwH represents an enhanced version of standard NSR.

\vspace{-0.3cm}

\begin{figure}[H]
\makebox[\textwidth][c]{
\begin{tabular}{cccc}
\includegraphics[scale = 0.24]{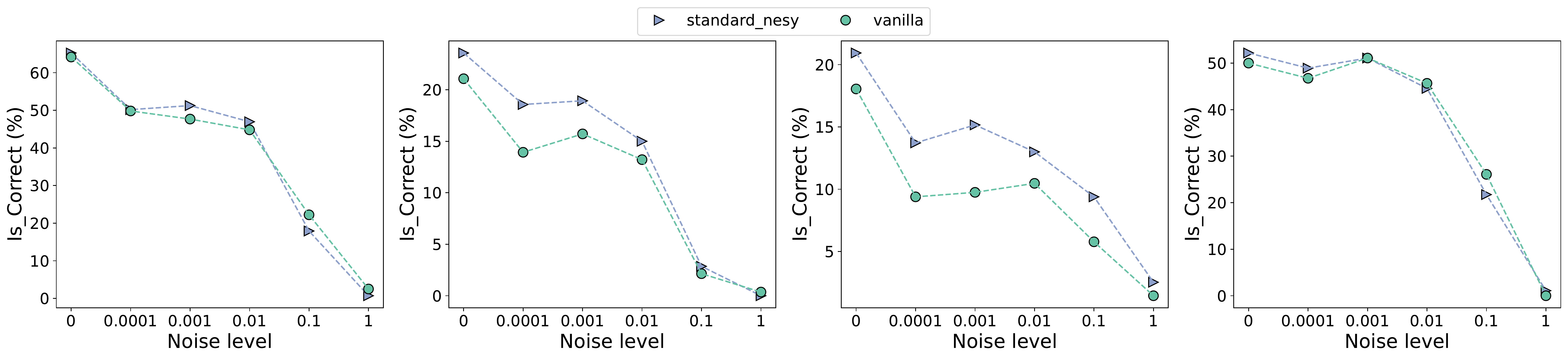}\\
\end{tabular}}
\caption{\textbf{Masked NSRwH vs. NeSymReS.} Comparison between fully masked NSRwH (\texttt{vanilla}) and standard NeSymReS (\texttt{standard$\_$nesy}) for different noise levels for the 
\texttt{train$\_$nc},
\texttt{train$\_$wc}, \texttt{only$\_$five$\_$variables$\_$nc} and \texttt{AIF} datasets from left to right.}\label{comp2}
\end{figure}

\begin{figure}[H]
\makebox[\textwidth][c]{
\begin{tabular}{cccc}
\includegraphics[scale = 0.24]{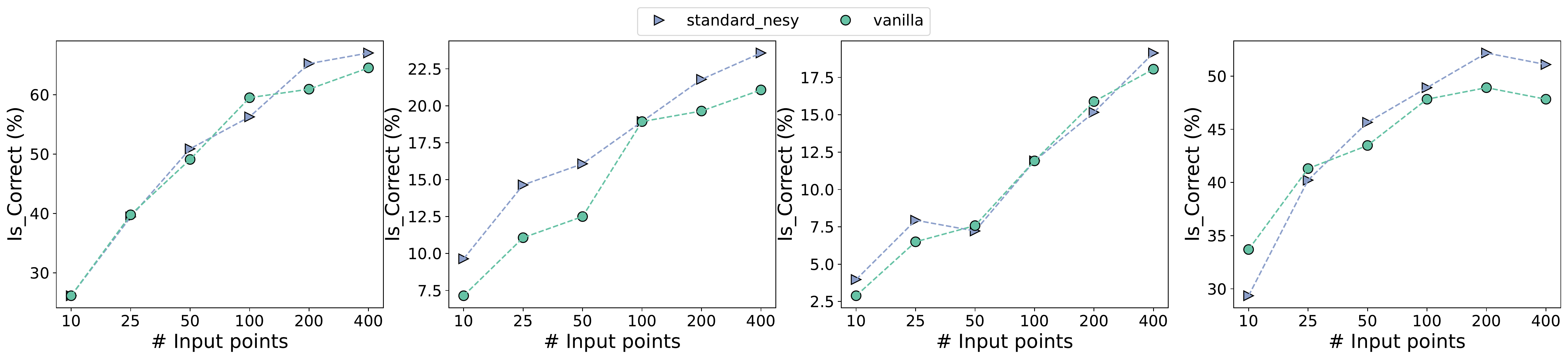}\\
\end{tabular}}
\caption{\textbf{Masked NSRwH vs. NeSymReS.} Comparison between fully masked NSRwH (\texttt{vanilla}) and standard NeSymReS (\texttt{standard$\_$nesy}) for a different number of input points for the 
\texttt{train$\_$nc},
\texttt{train$\_$wc}, \texttt{only$\_$five$\_$variables$\_$nc} and \texttt{AIF} datasets from left to right.}\label{comp1}
\end{figure}

\end{document}